\documentclass[journal]{IEEEtran}
\usepackage{amsmath,amsfonts}
\usepackage{algorithmic}
\usepackage{algorithm}
\usepackage{array}
\usepackage[caption=false,font=normalsize,labelfont=sf,textfont=sf]{subfig}
\usepackage{textcomp}
\usepackage{stfloats}
\usepackage{url}
\usepackage{multirow} 
\usepackage{verbatim}
\usepackage{graphicx}
\usepackage{cite}
\usepackage{tikz,xcolor}
\usepackage[implicit=false]{hyperref}
 
\hypersetup{hidelinks,
	colorlinks=true,
	allcolors=black,
	pdfstartview=Fit,
	breaklinks=true}
	
\definecolor{lime}{HTML}{A6CE39}
\DeclareRobustCommand{\orcidicon}{
\begin{tikzpicture}
\draw[lime, fill=lime] (0,0)
circle[radius=0.13]
node[white]{{\fontfamily{qag}\selectfont \tiny \.{I}D}};
\end{tikzpicture}
\hspace{-2mm}
}
\foreach \x in {A, ..., Z}{%
\expandafter\xdef\csname orcid\x\endcsname{\noexpand\href{https://orcid.org/\csname orcidauthor\x\endcsname}{\noexpand\orcidicon}}
}
 
\begin{document}

\title{Generating Event-oriented Attribution for Movies via Two-Stage Prefix-Enhanced Multimodal LLM}

\author{Yuanjie Lyu\orcidA{}\hspace{-1mm}, Tong Xu\orcidB{}\hspace{-1mm}, Zihan Niu, Bo Peng, Jing ke, Enhong Chen\orcidC{}\hspace{-1mm}, ~\IEEEmembership{Fellow, IEEE}

\thanks{Y. Lyu, T. Xu, Z. Niu, B. Peng, J. Ke,  and E. Chen are with the Anhui Province Key Lab of Big Data Analysis and Application, School of Computer Science, University of Science and Technology of China, Hefei 230026, China (e-mail: s1583050085@gmail.com, tongxu@ustc.edu.cn, niuzihan@mail.ustc.edu.cn, pb1150300625@mail.ustc.edu.cn, kejing@mail.ustc.edu.cn, cheneh@ustc.edu.cn)}
\thanks{Manuscript received XXX XX, 2024.}
}

\markboth{IEEE TRANSACTIONS ON XX, VOL. XX, NO. XX, XX. 2024}%
{Shell \MakeLowercase{\textit{Yuanjie Lyu et al.}}: A Sample Article Using IEEEtran.cls for IEEE Journals}


\maketitle

\begin{abstract}
The prosperity of social media platforms has raised the urgent demand for semantic-rich services, e.g., event and storyline attribution. However, most existing research focuses on clip-level event understanding, primarily through basic captioning tasks, without analyzing the causes of events across an entire movie. This is a significant challenge, as even advanced multimodal large language models (MLLMs) struggle with extensive multimodal information due to limited context length. To address this issue, we propose a Two-Stage Prefix-Enhanced MLLM (TSPE) approach for event attribution, i.e., connecting associated events with their causal semantics, in movie videos. In the local stage, we introduce an interaction-aware prefix that guides the model to focus on the relevant multimodal information within a single clip, briefly summarizing the single event. Correspondingly, in the global stage, we strengthen the connections between associated events using an inferential knowledge graph, and design an event-aware prefix that directs the model to focus on associated events rather than all preceding clips, resulting in accurate event attribution. Comprehensive evaluations of two real-world datasets demonstrate that our framework outperforms state-of-the-art methods.

\end{abstract}


\section{Introduction}
\begin{figure*}[t]
    \centering
    \includegraphics[width=1.0\textwidth]{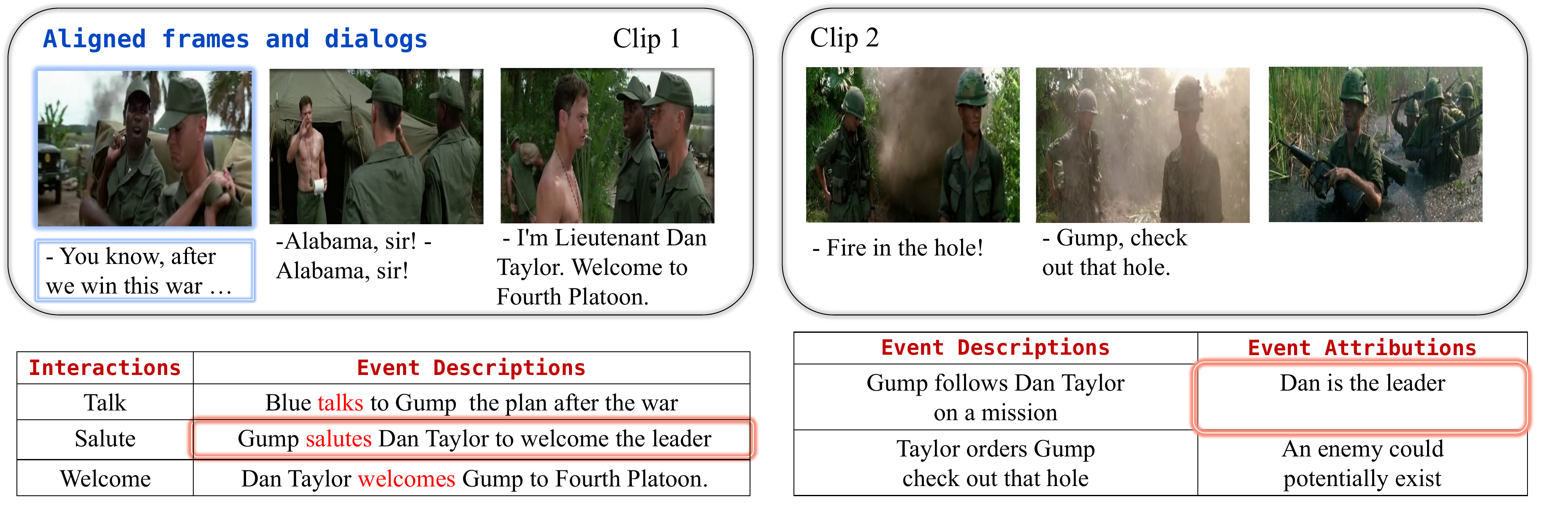}
    \caption{
    A toy example selected from the MovieGraph dataset, which contains two clips from the movie \emph{Forrest Gump}. Specifically, Clip 1 contains three events, and Clip 2 contains two events. The reason behind the event "Gump follows Dan Taylor on a mission" in Clip 2 is derived from the event "Gump salutes Dan Taylor" in Clip 1.
    }
    \label{fig:intro}
\end{figure*}

With the booming of social media platforms, viewing movies and TV dramas online has become a crucial part of daily entertainment, which raises the urgent demand for semantic-rich applications to provide event-oriented intelligent services~\cite{1318644}. Among them, the event attribution task, which targets connecting the associated video clips via their causal semantics, has been widely treated as a crucial tool for users to enhance their experience. For instance, users may read the plot synopsis first to select videos of interest, and further refer to the causal hints of the storyline to understand the event sequence, especially when viewing long dramas like Game of Thrones~\cite{10233916}. Unfortunately, existing research focuses on understanding events at the clip level, mainly through simple captioning tasks~\cite{lin2022swinbert}. So there is a notable lack of studies examining why events occur over the entire span of a movie (about 2 hours). This is a challenging task, even with Multimodal Large Language Models (MLLMs)~\cite{chen2023videollm, zhang2023video, Dai2023InstructBLIPTG}, it is challenging to integrate multimodal cues and interpret the underlying semantic relationships in long videos, due to the limited context length. Additionally, the numerous, overlapping, and complex events in a movie further complicate this understanding.


In this paper, we aim to understand the reasons behind each event from the perspective of the entire movie. Evidently, it is impractical to input an entire movie into an MLLM due to limited context length. A more feasible approach is to segment the movie into clips and process them in two stages: local understanding and global analysis. In the \textbf{local stage}, we comprehend the events within each clip, briefly summarizing the single event. In the \textbf{global stage}, we analyze the reasons for these events from the perspective of the entire movie. When determining the cause of an event in a specific clip, only the relevant clips need to be considered, not all preceding ones. However, this approach still introduces two challenges:
\begin{itemize}
    \item \textbf{Local Stage Challenge}. Ideally, each clip should contain a single event to simplify cause analysis. However, most segmentation methods divide clips based on scenes, often resulting in multiple events in a single clip. 
    For example, in Figure \ref{fig:intro}, Clip 1 contains 3 events, making it difficult to match the correct multimodal cues to each event. 

    \item \textbf{Global Stage Challenge}. Due to the large number of clips in a movie, inputting all preceding clips is impractical. Determining which clips are relevant is crucial. For instance, Clip 1 \& 2 in Figure \ref{fig:intro} are correlated, since Clip 1 provides the \emph{reason} (``leader”) for the \emph{events} (``follow on a mission”) in Clip 2. But, this correlation could not be captured by simply measuring the similarity between these two sentences. 
\end{itemize}


To address these challenges, this paper proposes a \textbf{T}wo-\textbf{S}tage \textbf{P}refix-\textbf{E}nhanced MLLM (TSPE) approach, designed to improve the performance of existing MLLMs in both stages.
In the \textbf{local stage}, we develop an interaction-aware prefix, which ensures the MLLM focuses on the relevant multimodal cues while minimizing interference from other overlapping events within the same clip. As illustrated on the left side of Figure \ref{fig:intro}, each event is tied to specific social interactions, which can be summarized by short phrases like “salute” and “welcome”. These interactions can be easily extracted using prompt-based MLLMs or existing recognition techniques\cite{kukleva2020learning,wu2021linking,qin2023when}. To leverage this, we designed an interaction-aware prefix using an attention mechanism, where social interaction serves as the query and multimodal cues act as the key and value, compressing information related to specific interactions into embeddings. These embeddings serve as prefixes for the MLLM, guiding it to focus on multimodal information relevant to the corresponding events and interactions in a single clip, while reducing interference from overlapping events. During training, we fine-tune both the model and the prefixes to briefly summarize the events.


After summarizing the single event in the local stage, we shift our focus to exploring relationships between events across the entire movie in the \textbf{global stage}. We also develop an event-aware prefix, which ensures the model focuses on relevant preceding events, rather than all prior events. However, traditional attention mechanisms may struggle to compress relevant events due to low semantic similarity between associated events. 
To address this, we incorporate an external knowledge graph, ATOMIC~\cite{sap2019atomic}, designed to predict potential outcomes based on previous events. We fine-tune the FlanT5 language model on ATOMIC, enabling it to predict possible consequences for any given event. Using ATOMIC's predictions as contextual information, we enhance semantic connections between events. When analyzing the cause of a current event, we gather all preceding events along with additional information from ATOMIC, treating them as key-value pairs while treating the current event as the query. We use an attention mechanism to compress these relevant events into embeddings, termed event-aware prefixes. These prefixes are then fed into the MLLM, guiding it to focus on information pertinent to event causes. During training, we fine-tune both the model and the prefixes for more accurate event attribution.

In our experiments, we evaluate our model using the MovieGraph~\cite{moviegraphs} dataset, a primary source of fine-grained events with attribution annotations. Additionally, we developed a new self-constructed dataset called CHAR (Character Behavior Analysis and Reasoning). We conduct comparative experiments and ablation experiments on both datasets to verify the effectiveness of our model\footnote{All codes and datasets will be made publicly available upon the paper's acceptance.}.

To the best of our knowledge, the technical contribution of this paper could be summarized as follows:

\begin{itemize}
    \item We are among the first ones who explore stories in long videos from the perspective of event attribution, which enhances the semantic level of video understanding.
    \item A novel two-stage framework is proposed with the Prefix-Enhanced MLLM, which effectively captures, summarizes, and enriches the multi-modal cues.
    \item Extensive evaluations of real-world datasets demonstrated the effectiveness of our solution compared with SOTA baseline methods
\end{itemize}

\section{Related Work}
Video understanding technology has progressed to automatically extract high-level semantic information (e.g., interactions, social relationships, events) from videos. These semantics are vital for comprehending video stories. Moreover, visual-language pre-training models allow us to interpret this high-level semantic information and leverage it for text generation. These technologies create the possibility for machines to autonomously comprehend and generate stories from videos. 
\subsection{Video Semantic Understanding}
Video understanding involves multiple levels of semantics. Low-level semantic video understanding attempts to identify the basic elements of videos, such as actions~\cite{simonyan2014two,wang2015action} and objects~\cite{ren2015faster,he2017mask}, which has achieved significant progress and maturity. 
Correspondingly, semantic-level understanding for video stories requires analyzing semantic factors beyond actions and objects, such as social relationships~\cite{Liu_2019_CVPR, wu2021linking, qin2023when}, interactions~\cite{kukleva2020learning}, movie summaries~\cite{10195243, zhou2019character} and events~\cite{martin2018event}. 
These factors are more difficult to identify, which have attracted wide research interest in recent years. Along this line, prior arts like MovieGraphs~\cite{moviegraphs}, MovieNet~\cite{huang2020movienet} 
provided rich annotations of semantic-level cues, and supported various downstream video understanding tasks. 





Generally, though semantic-level cues in videos are crucial for story understanding, most existing works only identify this information on short-term video clips, ignoring the fact that a complete story often spans a longer time period and more complex logical relationships. Meanwhile, only a few works have attempted to explore this issue from a long-term perspective. For example, 
Mittal et al.~\cite{mittal2021affect2mm} use emotion causation theories to model the emotional state evoked in movie clips. Wu et al.~\cite{wu2022scene} divide the scenes of movies according to the storyline. Lin et al.~\cite{lin2022makes} propose a novel framework that integrates two types of commonsense knowledge for future event generation. Wu and Krahenbuhl ~\cite{wu2021towards} introduce a Long-form Video Understanding (LVU) benchmark with multiple long-form tasks. Different from prior arts, in this paper, we mainly focus on the semantic-level event description and attribution task for long videos.

\subsection{Vision-Language Pre-trained Model}
Visual language pre-training aims to learn a multi-modal foundation model that can enhance the performance of various visual and language tasks. These tasks can be either understanding-based (e.g. image-text retrieval) or generation-based (e.g. image captioning). Different downstream tasks require different model architectures, including dual-encoder architecture~\cite{radford2021learning,jia2021scaling}, fused-encoder architecture~\cite{li2021align,Tan2019LXMERTLC}, encoder-decoder architecture~\cite{li2023blip,li2022blip}, etc. 

Recently, as LLMs showed remarkable ability in comprehension and reasoning, many works tried to build VLP based on LLMs, such as MiniGPT-4~\cite{zhu2023minigpt}, Video-ChatGPT~\cite{li2023videochat}, VideoChat~\cite{maaz2023video}, Instruct-BLIP~\cite{Dai2023InstructBLIPTG} and Video-LLaMA~\cite{zhang2023video}. Generally, all of these works consist of a visual encoder, a projection layer that aligns visual and text tokens and a pretrained LLM as the backbone. These attempts achieve good performance on many visual tasks like action detection and visual question answering. 
However, these models can't achieve satisfying performance anymore when facing long and complex videos. These models lack ability to filter out effective multi-modal cues from so much irrelevant noise.



Although directly utilizing VLPs can't work well in long and complex videos, they can serve as good foundations to support more fine-grained operations. Some recent works~\cite{lin2022revive,li2023decap,shao2023prompting} have applied VLP to various domains and achieved significant performance improvements. Inspired by these works, we choose the BLIP-2 model as the backbone of our model, which can flexibly transfer to both vision-language understanding and generation tasks.

\section{Method}

\begin{figure*}[t]
    \centering
    \resizebox{\textwidth}{!}{\includegraphics{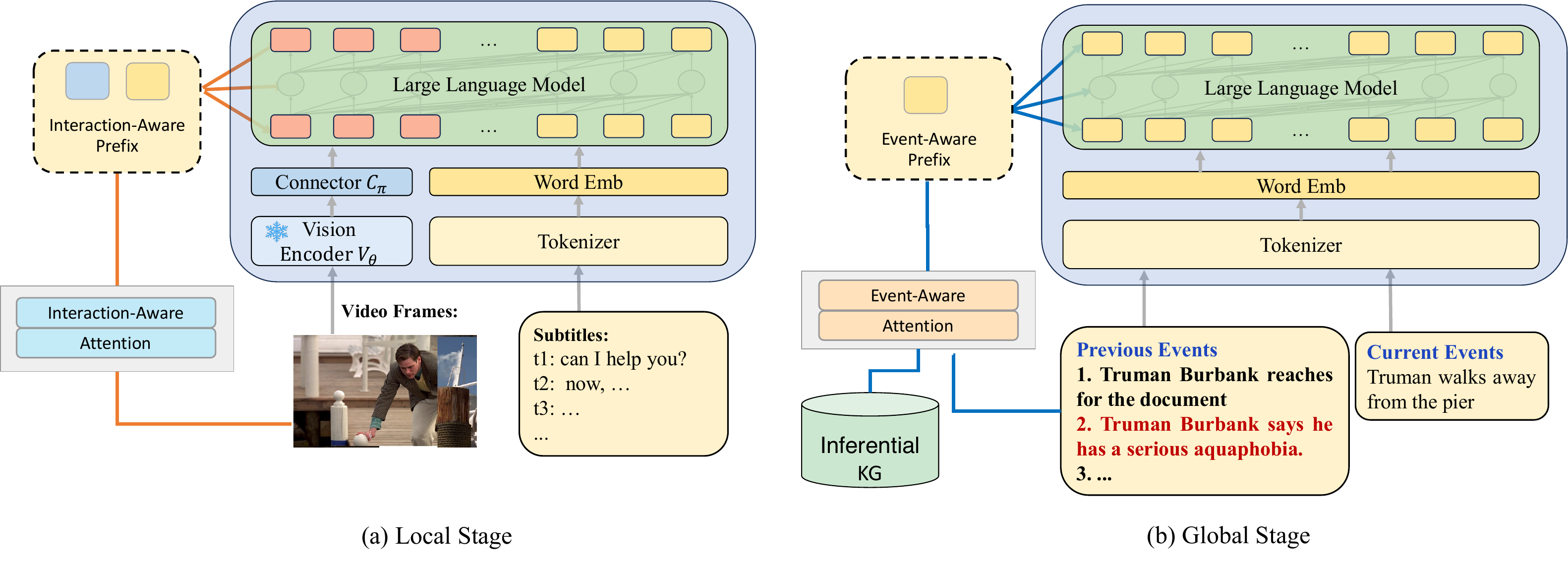}}
    \caption{
    Illustration of the Two-Stage Prefix-Enhanced MLLM (TSPE) method for event attribution in movie videos: Stage 1 extracts multimodal cues to briefly summarize the event, while Stage 2 uses these descriptions to infer underlying event causes. During training, only the last three layers of the LLM and the prefixes are fine-tuned.
    }
    \label{fig:framework}
\end{figure*}

We propose a Two-Stage Prefix-Enhanced MLLM (TSPE) approach for event attribution in movie videos, as illustrated in Figure ~\ref{fig:framework}. The method comprises two stages: local and global.

\subsection{Local Stage}
\begin{figure}[t]
    \centering
    \resizebox{0.5\textwidth}{!}{\includegraphics{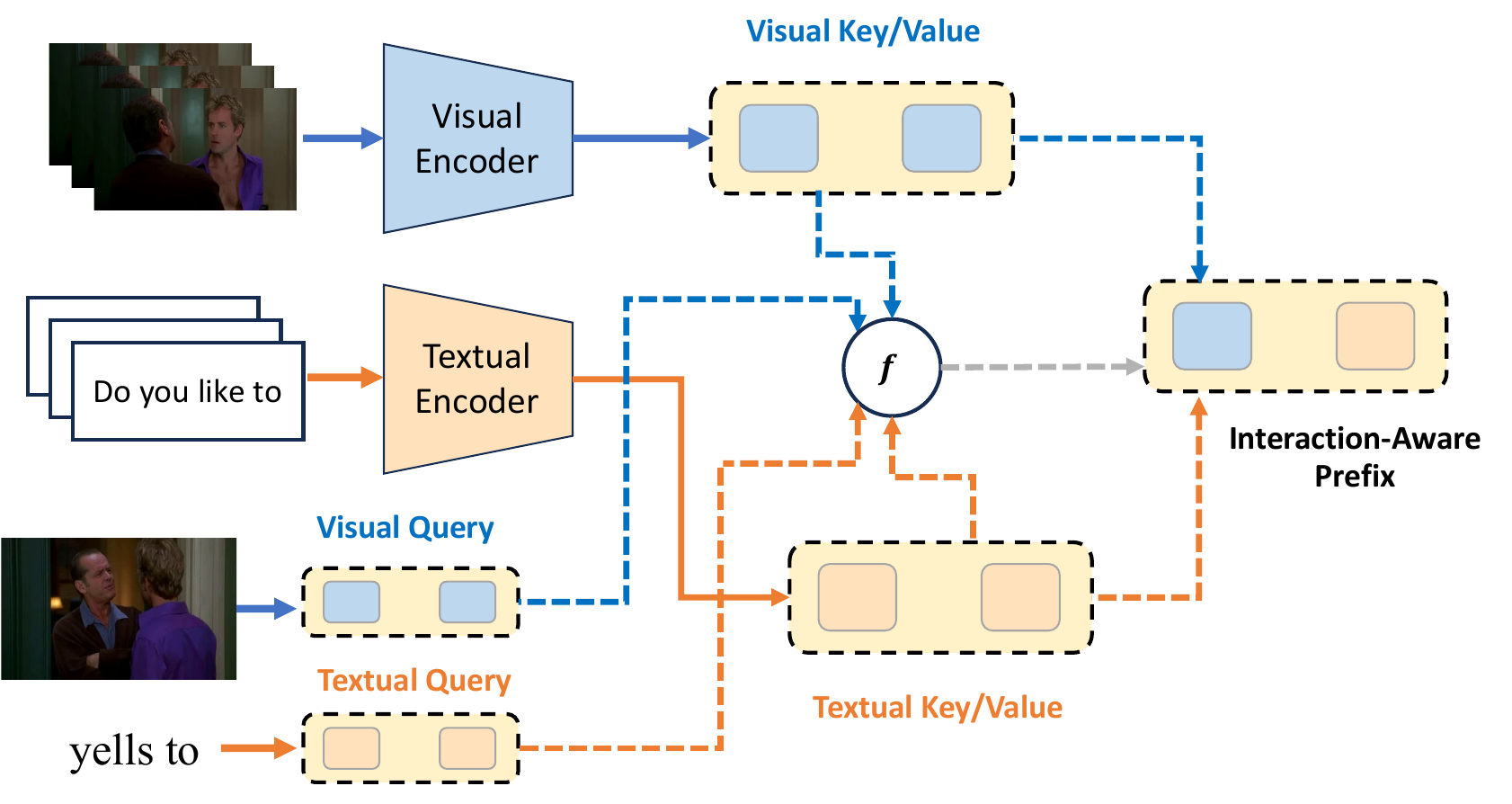}}
    \caption{
    In the local stage, we compress the textual and visual information from video frames and subtitles into event-related embeddings. An attention mechanism then measures the semantic relevance between the content and interactions, creating an interaction-aware prefix. This prefix is used as input to the LLM, which is fine-tuned to summarize the event in a single clip.
    }
    \label{fig:inter}
\end{figure}

In the local stage, the goal is to extract multimodal cues from a clip and convert them into high-level event summarization, facilitating subsequent event attribution. To achieve this, as Figure \ref{fig:inter} displays, we compress the complex multimodal cues within the clip into event-related embeddings, and use them to guide the model focus on multimodal cues relevant to the corresponding events while minimizing interference from other overlapping events. Specifically, an attention mechanism is employed to measure the semantic relevance between the content of frames and subtitles and the given social interaction, and we then fuse them into an embedding called the interaction-aware prefix. This prefix is used as the input prefix for the LLM.

In the following sections, we will introduce it in detail.



\textbf{Preprocessing.}
Given that narrative videos are often lengthy, the first step is typically to segment them into shorter clips by scene. As segmentation is not the focus of our research, we assume that the videos have already been segmented by scene, with each clip containing one or more complete events.

In our approach, we apply the BLIP-2 as the MLLM, which consists of 3 components: a pre-trained image transformer~\cite{dosovitskiy2020image} $F_v$ for visual feature extraction, a text transformer FlanT5~\cite{chung2022scaling}  $F_s$ that can function as both a text encoder and a text decoder, and a lightweight Q-former that effectively bridges the modality gap. We uniformly sample $t$ frames $\{v_1, v_2, ..., v_t\}$ from the video clip by the frequency as 1 frame/second, with subtitle text  $\{s_1, s_2, ..., s_t\}$ temporally aligned with frames. 
Then obtain frame feature $f_{v_i}$ as $f_{v_i} = F_v(v_i)$. We subsequently apply a Q-former to these frame features to bridge the modality gap, resulting in visual query features.
Finally, given the visual query features and the subtitle features, a straightforward approach to summarize events is to feed all these elements to the BLIP-2.

\textbf{Interaction-Aware Prefix.}
As mentioned earlier, a straightforward way to summarize events in the local stage is to input all multimodal cues into BLIP-2. However, the visual frames and textual subtitles in video clips often contain dense information that is irrelevant to the event. To capture the most relevant details for events, we propose the \emph{Interaction-Aware Prefix-Enhance} mechanism. As Figure \ref{fig:inter} shows, this mechanism assigns varying weights to short-term subtitles and frames based on their relevance to the event, and then fuses them into an embedding. This embedding will guide the MLLM in converting multimodal cues into high-level event summarization. 

We implement this method by analyzing social interactions. Since the core of events is the social interactions among characters, which can be detected by existing methods~\cite{kukleva2020learning}, we measure the relevance of multi-modal cues to social interactions, which is equivalent to measuring the relevance of the corresponding events.

Clearly, if a short-term subtitle, denoted as $s_i$, is related to interaction $I$, then they should exhibit a high degree of semantic similarity. Building upon this, we construct the text portion of the interaction-aware attention mechanism. 
We leverage the encoder of the powerful pre-trained Flan-T5~\cite{chung2022scaling}, for extracting features from both subtitle and interaction text. 
And during the training process, we freeze it:
\begin{equation}
    H_{s_i} = F_s( s_i ), H_{I_s} = F_s(I_s). 
    \label{eq: Fs}
\end{equation}
To obtain the global representations of the subtitle and interaction, we attach a [CLS] token to the beginning of both $s_i$ and $I_s$, utilizing the hidden states of the [CLS] tokens, denoted as $h_{s_i}$ and $h_{I_s}$, as the global representations for them.

We take the subtitle representations from time step $1$ to $t$ ($\{h_{s_{1}}, h_{s_{2}}, ...,h_{s_{t}}\}$) as the key and the interaction representation ($h_{I_s}$) as the query, and employ the dot-product attention mechanism~\cite{vaswani2017attention} to calculate the score indicating the degree of association between short-term subtitles and the interaction: 
\begin{equation}
   A_s = \mathrm{Softmax}(\frac{\mathbf{W_q} h_{I_s} (\mathbf{W_k} \{h_{s_{1}}, h_{s_{2}}, ... ,h_{s_{t}}\})^T}{\sqrt{d}}),
   \label{eq: A_s}
\end{equation}
where $d$ represents feature dimension and $\mathbf{W_q}, \mathbf{W_k}$ correspond to the learnable parameters.

Similar to the text portion, we also need to filter out the irrelevant visual elements for the event. We feed the image depicting the interaction
between two individuals 
into the image transformer to extract the visual features of the interaction, denoted as $f_{I_v}$. Additionally, we also employ a dot-product attention mechanism to compute the attention scores that quantify the correlation between frame features 
and the interaction: 
\begin{equation}
   A_v = \mathrm{Softmax}(\frac{\mathbf{W_q} f_{I_v} (\mathbf{W_k} \{f_{v_{1}}, f_{v_{2}}, ... ,f_{v_{t}}\})^T}{\sqrt{d}}).
   \label{eq: A_v}
\end{equation}

Since subtitles are temporally aligned with frames, we aggregate and normalize the attention scores from both visual and textual sources, denoted as $\mathbf{A} = \mathrm{normalization}(A_s + A_v)$. 
$\mathbf{A}$ reflects the degree of association between multi-modal information and interactions, i.e. the degree of association between multi-modal information and events.

Then, we leverage the attention score matrix $\mathbf{A}$, to weigh the frame features and subtitle features. This operation results in interaction-aware visual and textual context. Essentially, it highlights the relevant visual frames and corresponding subtitles that contribute to the understanding of the event.
\begin{align}
   H_v & = \mathbf{A} \{f_{v_1}, f_{v_2}, \ldots ,f_{v_t}\}  \mathbf{W_v}, \\
   H_s & = \mathbf{A} \{h_{s_1}, h_{s_2}, \ldots ,h_{s_t}\}  \mathbf{W_v},
\end{align}
where $\mathbf{W_v}$ represents the learnable parameters. To bridge the modality gap, we feed the interaction-aware visual context $H_v$ into the Q-former. 


While the interaction-aware context effectively aggregates the most relevant information at each time step, multimodal information often spans multiple time steps. Relying solely on the interaction-aware context risks missing important cues. To address this, we use the interaction-aware context as prefixes and feed the complete multimodal information into the MLLM. This hierarchical structure allows for better capture of the most relevant multimodal cues across multiple time steps.

Specifically, when generating the $i$-th token of the description, we directly concatenate the interaction-aware prefix with the previous word embedding, and then feed them to the decoder. Simultaneously, we concatenate all subtitle features and frame features from time 1 to time $t$ as the full textual and visual information to the decoder. The attention mechanism in the decoder will be further adapted to capture the most relevant multi-modal cues through interaction-aware prefixes:
\begin{equation}
   v_i = \mathrm{DEC}(w_{1:i-1} \oplus H_v \oplus H_s, H_{s_{1}:s_{t}},  H_{v_{1}:v_{t}}).
\end{equation}
Here, $H_s$ and $H_v$ represent the salient information for understanding events, $H_{s_{1}:s_{t}}$ represents the full textual information, and $H_{v_{1}:v_{t}}$ represents the full visual features obtained by processing frame features through the Q-former. The variable $w_{1:i-1}$ denotes the previous tokens that have been generated.

Then, we use linear layers and a softmax layer to classify the feature $v_i$ output by decoder and compute the probability distribution of the $i$-th token.


\begin{figure*}[t]
    \centering
    \resizebox{0.8\textwidth}{!}{\includegraphics{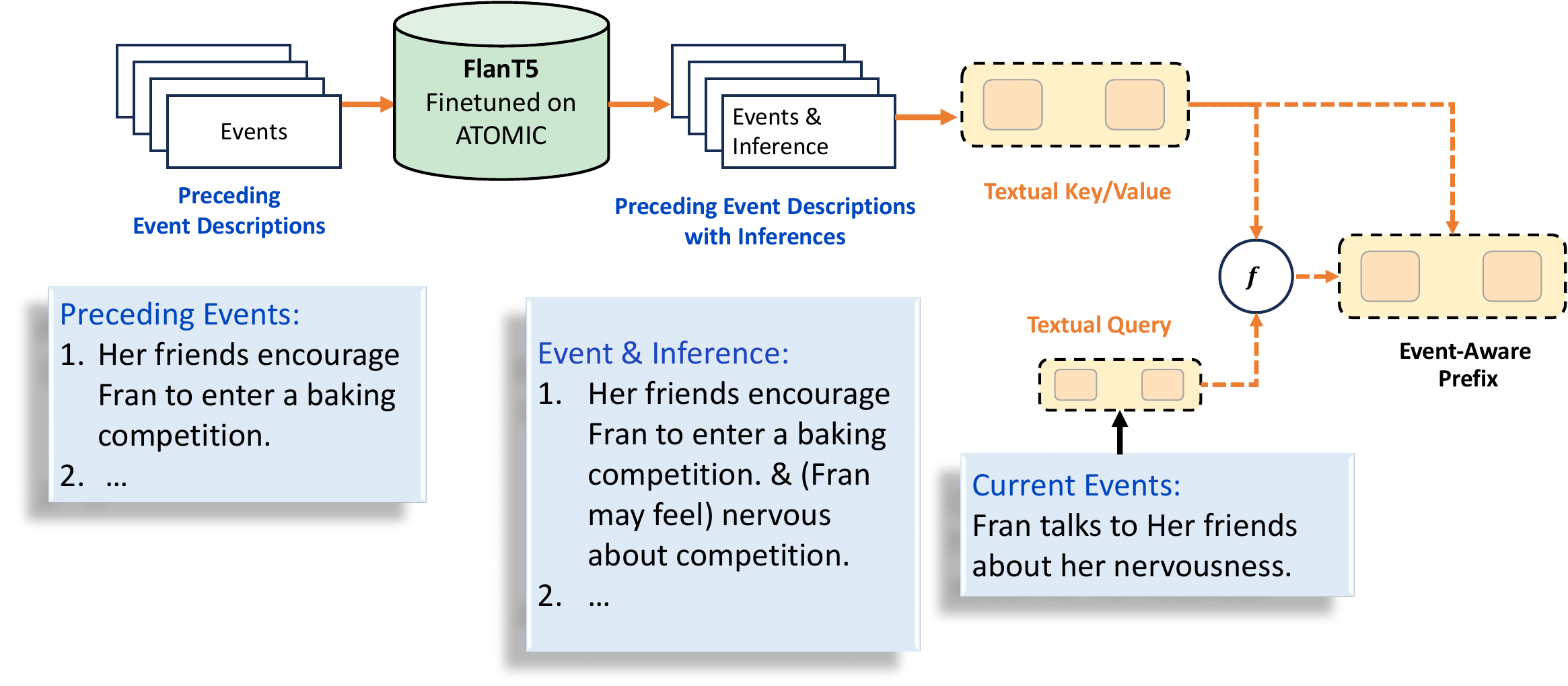}}
    \caption{
In the global stage, the results from the local stage are fed into a common sense knowledge graph, ATOMIC, to predict potential outcomes of prior events. An attention mechanism assesses the relevance between the current and previous events, fusing them into event-aware embeddings. These embeddings are then input into the LLM, which is fine-tuned to generate the underlying causes of the events.    
}
    \label{fig:global}
\end{figure*}

\subsection{Global Stage}

In the first stage, we briefly summarize events in the single clips, which are essential for understanding the stories of videos. However, this step only captures the story that occurs at a specific moment, while a coherent story is composed of many logically connected events.
Therefore, we need to explore the underlying reasons behind the events and comprehend the story at the whole video level.

In the second stage, we further generate the reasons for each event based on the summarization of the event as shown in Figure \ref{fig:global}. A key challenge is that while the current event is influenced by previous events, not all prior events are relevant. To address this issue, we input all previous event descriptions into an inferential common sense knowledge graph, ATOMIC, to predict the potential outcomes of prior events, providing additional context. Next, an attention mechanism similar to that in the previous stage is employed to measure the semantic relevance between the current event and previous events. Finally, the previous events are fused into embeddings based on attention scores, forming what we call the event-aware prefix. This prefix is then input into the LLM, and it will help the model focus on information pertinent to event causes.

In the following sections, we will introduce it in detail.


\textbf{Inferential Knowledge Graph.} 
Considering the current event description often has significant semantic differences from associated previous events, we propose incorporating an inferential knowledge graph to address this issue. This approach leverages commonsense reasoning to infer potential outcomes of the event, including its causes, effects on one character, and effects on others.

As shown in Figure \ref{fig:global}, we present an example of how external knowledge can enhance the semantic similarity between events. The preceding event is that Fran’s friends encourage Fran to enter a baking competition. The current event is that Fran talks to her friends about her nervousness. These two events are connected but semantically dissimilar, which will invalidate the attention mechanism based on semantic similarity. However, by using the Inferential KG, we can add information to the preceding event: Fran may feel nervous about the competition. This makes the preceding event more semantically similar to the current event. 

To infer commonsense knowledge related to the events, we leverage ATOMIC~\cite{sap2019atomic}, a large-scale commonsense KG with 9 inference relations. Although ATOMIC contains commonsense knowledge about many events, it can not cover all the events. To address this issue, inspired by Bosselut et al. ~\cite{bosselut-etal-2019-comet}; Lin et al. ~\cite{lin2022makes}, we can expand the implicit knowledge in pre-trained language models into explicit knowledge in the commonsense KG. 
Specifically, we fine-tune FlanT5, a pre-trained language model, on ATOMIC to generate textual descriptions of commonsense knowledge for unseen events. 

Table \ref{tab:atomic} shows an example of ATOMIC. When fine-tuning the FlanT5 on ATOMIC, given an event “PersonX repels PersonY’s attack” and a relation “xIntent” in ATOMIC, the model was trained to generate “to protect others” as the output. Similarly, given the same event and the relation “xEffect”, the model was trained to generate “heart races” as the output. When using the fine-tuned model, given an event that is not present in ATOMIC, such as Fran’s friends encourage Fran to enter a baking competition, we anonymize the named entities and adjust it to the event format on ATOMIC: PersonX encourages PersonY to enter a baking competition. Then, we input the relation we want to know, such as OReact (representing PersonY’s possible reaction), and FlanT5 generates sentences, such as PersonY may feel nervous. This way, we can generate commonsense knowledge for any event, even those unseen by the model.

The fine-tuned model leverages the implicit knowledge of the pre-trained language model to handle some events that are not explicitly covered by ATOMIC. We generate 9 types of event-related inference texts for each event, and concatenate them directly with the event as additional context. 

\label{sec:atomic}
\begin{table}[t]
    \centering
    \caption{An example in ATOMIC.}
\resizebox{0.45\textwidth}{!}{
\begin{tabular}{|c|c|c|}
\hline \multicolumn{3}{|c|}{ Input Event: PersonX repels PersonY's attack } \\
\hline $\begin{array}{l}\text { xIntent } \\
\text { (PersonX intent) } \\
\text { to protect others }\end{array}$ & $\begin{array}{l}\text { xEffect } \\
\text { (PersonX effect) } \\
\text { heart races }\end{array}$ & $\begin{array}{l}\text { oReact } \\
\text { (Other react) } \\
\text { weak; ashamed }\end{array}$ \\
\hline $\begin{array}{l}\text { xNeed } \\
\text { (PersonX need) } \\
\text { to defense himself }\end{array}$ & $\begin{array}{l}\text { xWant } \\
\text { (PersonX want to) } \\
\text { file a police report }\end{array}$ & $\begin{array}{l}\text { oWant } \\
\text { (Other want to) } \\
\text { attack again }\end{array}$ \\
\hline $\begin{array}{l}\text { xAttr } \\
\text { (PersonX attribute) } \\
\text { strong; brave }\end{array}$ & $\begin{array}{l}\text { xReact } \\
\text { (PersonX react) } \\
\text { angry; tired }\end{array}$ & $\begin{array}{l}\text { oEffect } \\
\text { (Other effect) } \\
\text { falls back }\end{array}$ \\
\hline
\end{tabular}
}
\label{tab:atomic}
\end{table}


\textbf{Event-Aware Prefix.}
After enriching the event descriptions with additional ATOMIC predictions, we use the event-aware mechanism to further refine the selection of previous events, and generate the reason behind the current event. We first add all the extracted inferential knowledge to the corresponding event descriptions, then use Flan-T5 as a feature extractor for events with inferential knowledge, extracting the features of previous events and the features of the current event.

We take the previous event representations from $1$ to $i-1$ $(\{h_{E_{1}}, h_{E_{2}}, ..., h_{E_{i-1}}\})$ as the key and the current event representation$(h_{E_{i}})$ as the query, and employ the dot-product attention mechanism to calculate the score indicating the degree of association between them. 
\begin{equation}
   A_E = \mathrm{Softmax}(\frac{\mathbf{W_q} h_{E_{i}} (\mathbf{W_k} \{h_{E_{1}}, ... ,h_{E_{i-1}}\})^T}{\sqrt{d}}).
\end{equation}
Then we leverage the attention score matrix $\mathbf{A_E}$, to weigh the previous event features. This operation results in an event-aware context. 
\begin{align}
   H_E & = \mathbf{A_E} \{h_{E_1}, h_{E_{2}}, \ldots ,h_{E_{i-1}}\}  \mathbf{W_v},
\end{align}
The structure of Event-Aware Attention is very similar to that of Interaction-Aware Attention. However, we do not use visual information for two reasons: (1) The event descriptions already capture most of the relevant visual information. (2) Storytelling videos are usually very long, such as movies that last up to 2 hours, 
so we can not handle all the visual information.

Considering that the cause of the event might be present in the subtitle of the current clip, we also include the subtitle of the current clip as input to the model. When generating the $i$-th token of the reasons, we directly concatenate the event-aware prefix with the previous word embedding, and then feed them to the decoder.
We also provide the decoder with full previous event features, and additional features: subtitle features in the current clip.

\begin{equation}
   v_i = \mathrm{DEC}(w_{1:i-1} \oplus H_E, H_{E_{1}:E_{i-1}}, H_{s_{1}:s_{t}}).
\end{equation}
Here, $H_E$ represents the salient event-aware prefix, the sequence $H_{E_{1}:E_{i-1}}$ represents the full information of previous events, the sequence $H_{s_{1}:s_{t}}$ represents the subtitle information in the current clip, and the variable $w_{1:i-1}$ denotes the previous tokens that have been generated.

Then, we use linear layers and a softmax layer to classify the feature $v_i$ output by decoder and compute the probability distribution of the $i$-th token.

\section{Experiment}

\begin{table}[t]
    \centering
    \caption{Statistics of the MovieGraph dataset and our self-constructed CHAR dataset.}
        \begin{tabular}{c|c|c}
            \hline Datasets & Annotations & Number   \\
            
            \hline \multirow{5}{*}{MovieGraph} 
            & Characters   & 2.98/clip \\
            & Interactions & 38,872 \\
            & Events        & 20,238  \\
            & Reasons      & 13,288  \\
            & Avg. Duration& 1min/clip  \\
           
            \hline 
            \multirow{5}{*}{CHAR} 
            & Characters    & 3.01/clip \\
            & Interactions & 9,608 \\
            & Events      & 9,529  \\
            & Reasons       & 4,634 \\
            & Avg. Duration & 5min/clip   \\
        \hline
        \end{tabular}
    \label{tab: moviegraph_char}
\end{table}

\begin{table*}[t]
    \centering
    \caption{Comparisons with state-of-the-art models on MovieGraph and self-constructed CHAR dataset in Local Stage. Numbers in bold mean that the best result.
    }
    \resizebox{0.95\textwidth}{!}{
        \begin{tabular}{c|c|ccccccc}
            \hline Dataset & Model & BLEU-1 &  BLEU-2 &  BLEU-3 & BLEU-4 & METEOR & ROUGE-L  \\
            
            \hline \multirow{5}{*}{ MovieGraph } 
            & FlanT5      & 0.098 & 0.039 & 0.019 & 0.011 & 0.049 & 0.107  \\
            & BLIP-2     & 0.112 & 0.054 & 0.029 & 0.015 & 0.069 & 0.133  \\
            & FlanT5-IE      & 0.124 & 0.058 & 0.033 & 0.020 & 0.051 & 0.129  \\
            & BLIP-2-IE    & 0.169 & 0.092 & 0.056 & 0.037 & 0.070 & 0.175 \\
            & TSPE & \textbf{0.196} & \textbf{0.110} & \textbf{0.071} & \textbf{0.048} & \textbf{0.080} & \textbf{0.206}\\
            \hline 
            \multirow{5}{*}{ CHAR } 
            & FlanT5              & 0.098 & 0.033 & 0.017 &  0.011 & 0.036 & 0.109 \\
            & BLIP-2             & 0.106 & 0.040 & 0.024 & 0.017 & 0.043 & 0.110  \\
            & FlanT5-IE          & 0.088 & 0.035 & 0.021 & 0.014 & 0.039 & 0.105 \\
            & BLIP-2-IE              & 0.120  &  0.049 & 0.029 & 0.020 & 0.045 & 0.121\\
            & TSPE   & \textbf{0.131} & \textbf{0.061} & \textbf{0.037} & \textbf{0.026} & \textbf{0.056} & \textbf{0.139}  \\
        \hline
        \end{tabular}
    }
    \label{tab: topic}
\end{table*}

\begin{table*}[t]
    \centering
    \caption{The performance of TSPE when using different models as backbone ON MOVIEGRAPH DATASET.
    }
    \resizebox{0.9\textwidth}{!}{
    \begin{tabular}{c|ccccccc}
    \hline
    Model & BLEU-1 &  BLEU-2 &  BLEU-3 & BLEU-4 & METEOR & ROUGE-L   \\
    \hline
    BLIP-IE & 0.147 & 0.065 & 0.032 & 0.018 & 0.057 & 0.157  \\
    
    BLIP-2-IE & 0.169 & 0.092 & 0.056 & 0.037 & 0.070 & 0.175  \\
    
    Instruct-BLIP-IE & 0.185 & 0.103 & 0.062 & 0.040 & 0.076 & 0.189 \\
    
    TSPE(BLIP) & 0.161 & 0.072 & 0.037 & 0.020 & 0.058 & 0.167  \\
    
    TSPE(BLIP-2) & 0.196 &0.110 &0.071 & 0.048 &0.080 &0.206  \\
    
    TSPE(Instruct-BLIP) & 0.202 & 0.120 & 0.079 & 0.057 & 0.081 & 0.206  \\
    \hline
    \end{tabular}
    }
    \label{tab: backbone}
\end{table*}

\begin{table*}[t]
    \centering
    \caption{Comparisons with state-of-the-art models on MovieGraph and self-constructed CHAR dataset in Global Stage. Numbers in bold mean that the best result.
    }
    \resizebox{0.95\textwidth}{!}{
        \begin{tabular}{c|c|cccccc}
            \hline Dataset & Model & BLEU-1 &  BLEU-2 &  BLEU-3 & BLEU-4 & METEOR & ROUGE-L   \\
            
            \hline \multirow{3}{*}{ MovieGraph } 
            & FlanT5             & 0.128 & 0.052 & 0.026 & 0.013 & 0.052 & 0.145 \\ 
            & FlanT5-KE          & 0.143 & 0.062 & 0.034 & 0.020 & 0.055 & 0.149  \\
            & TSPE  & \textbf{0.150} & \textbf{0.070} & \textbf{0.039} & \textbf{0.024} & \textbf{0.063} & \textbf{0.168} \\
            
            \hline 
            \multirow{3}{*}{ CHAR } 
            & FlanT5              & 0.171 & 0.083 & 0.052 & 0.035 & 0.066 & 0.176  \\
            & FlanT5-KE           & 0.173 & 0.087 & 0.054 & 0.035 & 0.061 & 0.178  \\
            & TSPE      & \textbf{0.181} & \textbf{0.091} & \textbf{0.056} & \textbf{0.037} & \textbf{0.071} & \textbf{0.185}  \\
        \hline
        \end{tabular}
    }
    \label{tab: reason}
\end{table*}

\subsection{Datasets and Evaluation Metrics.}

We conducted our experiments on two datasets, whose statistics are summarized in Table \ref{tab: moviegraph_char}. 

Due to the complexity of annotating semantics in videos, obtaining suitable training data is a challenge. Movie videos offer a practical solution. Although some datasets exist for understanding movie videos~\cite{han2023autoad}, fine-grained event datasets with attribution annotations from a holistic movie perspective are still scarce. Currently, MovieGraph~\cite{moviegraphs} is the primary source, which provides detailed graphical annotations for 51 movies, segmented into 7,637 scenes and 38,872 interactions. Following Kukleva et al.~\cite{kukleva2020learning}, the dataset is partitioned into train (35 movies), validation (7 movies) and test (9 movies) splits. Note that events in the MovieGraph dataset are composed of two parts: interactions and topics. Take the event "Gump talks to Dan Taylor the plan after the war" as an example, MovieGraph first annotated the interaction: Gump talks to Dan Taylor, and then annotated the topic: the plan after the war. 


We also independently constructed a dataset named CHAR (Character Behavior Analysis and Reasoning), which comprises 21 movies, each with an average duration of 2 hours. When constructing CHAR, we followed the semantic annotation method of the MovieGraph dataset. Notably, we placed particular emphasis on capturing the causal relationships between events. Annotators were instructed to take into account previous events when annotating the reasons for current events. We use the train (12 movies), validation (4 movies) and test (5 movies) splits. 

Comprehensive experiments are conducted on the two datasets in terms of four automatic metrics including BLEU~\cite{papineni2002bleu}, METEOR~\cite{banerjee2005meteor}, ROUGE-L~\cite{rouge2004package}. These metrics calculate the similarities and relevance between the generated text and reference. Considering that the local stage and the global stage of the TSPE model are trained separately, we also evaluate these two stages separately. Notably, in the local stage, the interaction and name text are already included in the model input. However, these elements often make up a significant portion of the event description. To ensure accurate evaluation metrics, we will remove the interaction text from the generated output during the evaluation.

\subsection{Experiment Setup}

\textbf{Baselines.}
In the local stage, we use BLIP-2 as MLLM, which has achieved SOTA performance in various multi-modal understanding and generation tasks. Therefore, we take BLIP-2 as a baseline. We also compare with FlanT5, a plain text generation model. We aim to verify the importance of multi-modal information for our task. We extend BLIP-2 and FlanT5 to adapt our task in two settings: (1) Origin: We directly feed the frames and subtitles into the model. (2) Interaction-Enhance(IE): We take the social interactions as additional information in the prompt, and feed frames, dialogues, and social interaction to the model. We retrained these models on the corresponding datasets when evaluating.

As mentioned earlier, for attribution generation, we excluded visual information in our experiments to conserve computational resources. Therefore, we employed the text-only model, FlanT5, as our baseline. This model takes current clip dialogues, relevant previous event descriptions selected via prompts, and the language model FlanT5, along with the current event descriptions as input. We also introduced a knowledge-enhanced variant, FlanT5-KE, which incorporates knowledge from the inferential knowledge graph as additional event information. We retrained these models on the corresponding datasets when evaluating.

\textbf{Implementation Details.}
We initialize the local stage model with the parameters of BLIP-2~\cite{li2023blip} model, and the global stage model with the parameters of FlanT5-XL~\cite{chung2022scaling} model. We also fine-tune FlanT5-XL on the ATOMIC dataset to generate the causal situation after the events. To train our local stage model efficiently, we applied a parameter freezing strategy that fixed the weights of the visual transformers and the encoder of the text transformer. This prevented them from being updated by the gradient descent algorithm. We only allowed the last 5 layers of the decoder to be trainable, while the rest of the layers were frozen. We followed a similar approach for our global stage model, where we froze the encoder of the FlanT5 model and only trained the last 5 layers of the decoder. Considering that there may be too many preceding events, we used prompts to instruct FlanT5 to remove those that are clearly irrelevant before they are fed into our model. We evaluate our approach on two datasets (MovieGraph and CHAR), and we use the same experimental settings for all experiments: batch size of 4, AdamW optimizer with a learning rate of 1e-5, and weight decay of 0.01. We use top-k sampling with k=4 to generate events and reasons. We ensure that all baselines have the same hyper-parameters as our model.

\begin{table}[t] 
\centering
    \caption{Ablation experiments on MovieGraph dataset in local stage.
    }
\resizebox{0.45\textwidth}{!}
{ 
    \begin{tabular}{c|ccccccc} 
        \hline 
         Model  & BLEU-4 & METEOR & ROUGE-L  \\
         \hline 
             TSPE       & \textbf{0.048} & \textbf{0.080} & \textbf{0.206} \\
            W/O VA         & 0.044 & 0.080 & 0.201 \\
            W/O TA         & 0.045 & 0.078 & 0.198 \\
            W/O VA \& TA   & 0.037 & 0.070 & 0.175 \\
            \hline 
    \end{tabular}
}
    \label{tab: ablation_topic}
\end{table}

\begin{table}[t] 
\centering
    \caption{Ablation experiments on MovieGraph dataset in global stage.
    }
\resizebox{0.48\textwidth}{!}
{ 
    \begin{tabular}{c|ccccccc} 
        \hline 
         Model   & BLEU-4 & METEOR & ROUGE-L  \\
         \hline 
             TSPE             & \textbf{0.024} & \textbf{0.063} & \textbf{0.168} \\
            
             W/O Subtitle   & 0.011 & 0.048 & 0.147 \\
             W/O Inferential KG  & 0.020 & 0.059 & 0.163  \\
             W/O Previous Events   & 0.013 & 0.051 & 0.143 \\
             W/O Event-Aware Module  & 0.020 & 0.055 & 0.149 \\
        \hline
    \end{tabular}
}
    \label{tab: ablation_event}
\end{table}

\subsection{Comparison with SOTA Methods
on Automatic Metrics
}
Considering that the local stage and the global stage of the TSPE model are trained separately, we also evaluate these two stages separately.

\textbf{Local Stage.} Table \ref{tab: topic} presents the automatic evaluation results of our TSPE framework model and the baselines on the local stage. The table reveals the following insights:

\begin{itemize}
    \item Our model surpasses all the baselines on all the evaluation metrics by a large margin, indicating that our model can effectively generate events without being affected by irrelevant multi-modal information.
    \item The BLIP-2 model, which leverages both visual and textual modalities, achieves higher scores than the FlanT5 model, which only uses textual information. This confirms the importance of multi-modal information for understanding events. 
    
    \item Social interaction information enhances event understanding, as shown by the improved performance of BLIP-2-IE and FlanT5-IE over BLIP-2 and FlanT5. However, our TSPE model outperforms BLIP-2-IE, demonstrating that interaction-aware prefix is more effective than simple concatenation of social interaction with other features.

    \item Event description generation on the CHAR dataset presents greater challenges compared to the MovieGraph dataset. This is primarily due to the CHAR dataset segmenting movies into clips with an average duration of 5 minutes, whereas the MovieGraph dataset's clips are on average only 1 minute long. Consequently, CHAR clips typically encompass more interactions, making it inherently more challenging to generate descriptions for each event.  

\end{itemize}

Our framework takes the BLIP-2 as the backbone,  and we did not compare it with other multi-modal baselines because it would be unfair. However, more other video-based methods are worth exploring to see if the model still has good performance against them. So we conducted additional experiments in the event description stage, using different backbones (BLIP, Instruct-BLIP) for our method and comparing them with the original backbone. The experimental results are displayed in Table \ref{tab: backbone}. Our model achieves superior performance over the original backbone with different backbones.


\textbf{Global Stage.} Table \ref{tab: reason} reveals the effects of different factors on the model’s ability to generate reasons behind the events. We can draw the following conclusions from the table: 

\begin{itemize}
    \item Using inferential KG, which generates possible outcomes of events, FlanT5-KE performs better than FlanT5. This shows that inferring the causes and consequences of events helps generate event reasons.
    \item  Although FlanT5-KE takes the knowledge to enhance the event understanding, it still inputs all the possible relevant events to the model equally. TSPE model adopts an event-aware attention mechanism similar to interaction-aware attention, which can filter out the irrelevant events more effectively, and thus perform better.

    \item  The improvement of FlanT5-KE over FlanT5 is less significant than the improvement of TSPE over FlanT5-KE. This indicates that external knowledge can enhance the semantic similarity between related events and help the model filter out useful information, but not directly capture the reasons behind the events.

\end{itemize}


\subsection{Ablation Study}
To investigate the effectiveness of the proposed modules, ablative experiments are conducted in the MovieGraph dataset. 
We are mainly concerned about the following points: 

\textbf{The impact of interaction-aware module on the local stage.} Table \ref{tab: ablation_topic} further shows that both visual and textual attention functions in the Interaction-aware module are important for the model’s performance.
\begin{itemize}
    \item If we omit the visual attention function in Eq (\ref{eq: A_v}) (shown in W/O VA column), which quantifies the correlation between the visual interaction and frame features, the model’s performance deteriorates on all metrics.

    \item If we omit the visual attention function in Eq (\ref{eq: A_v}) (shown in W/O TA column), which quantifies the correlation between the textual interaction and the short-term dialogues (shown in W/O TA column), the model’s performance deteriorates on all metrics.

    \item If we remove the Interaction-aware module (shown in W/O VA \& TA column), the model reduces to BILP-2-IE, which performs much worse than our model with the Interaction-aware module. 

\end{itemize}

\textbf{The impact of different factors on the global stage.} 
Our model relies on three main sources of information to generate the reasons for the events: subtitles in the current clip, previous events, and inferential KG. Table \ref{tab: ablation_event} shows that removing any of these sources will degrade the performance of our model.

\begin{itemize}
    \item As shown in Table \ref{tab: ablation_event}, if the character dialogue information is removed, the event reason generation ability of the model will drop significantly. Therefore, we can infer that many event reasons can be directly derived from the current character dialogue, rather than necessarily related to the previous events.
    
    \item  Previous events provide the context and coherence for the current event. If the previous event information is removed, the performance of the model will drop significantly. This proves that the current event is logically connected to the previous events. 

    \item  The event-aware module filters out the irrelevant previous events. Table \ref{tab: ablation_event} shows that without the event-aware module, our model degrades to FlanT5-KE, and performs much worse. This indicates that the prompt guiding FlanT5 to output the potentially relevant events has a lot of noise, and must be further filtered by the event-aware module in a fine-grained manner.
    
    \item Inferential KG enriches the events with additional information. As expected, without inferential information, the performance of the model dropped slightly. Since there are often multiple possible future events, ATOMIC cannot capture all of them, so the benefit of inferential KG is not significant. 
\end{itemize}

\begin{figure}[t]
    \centering
    \includegraphics[width=0.5\textwidth]{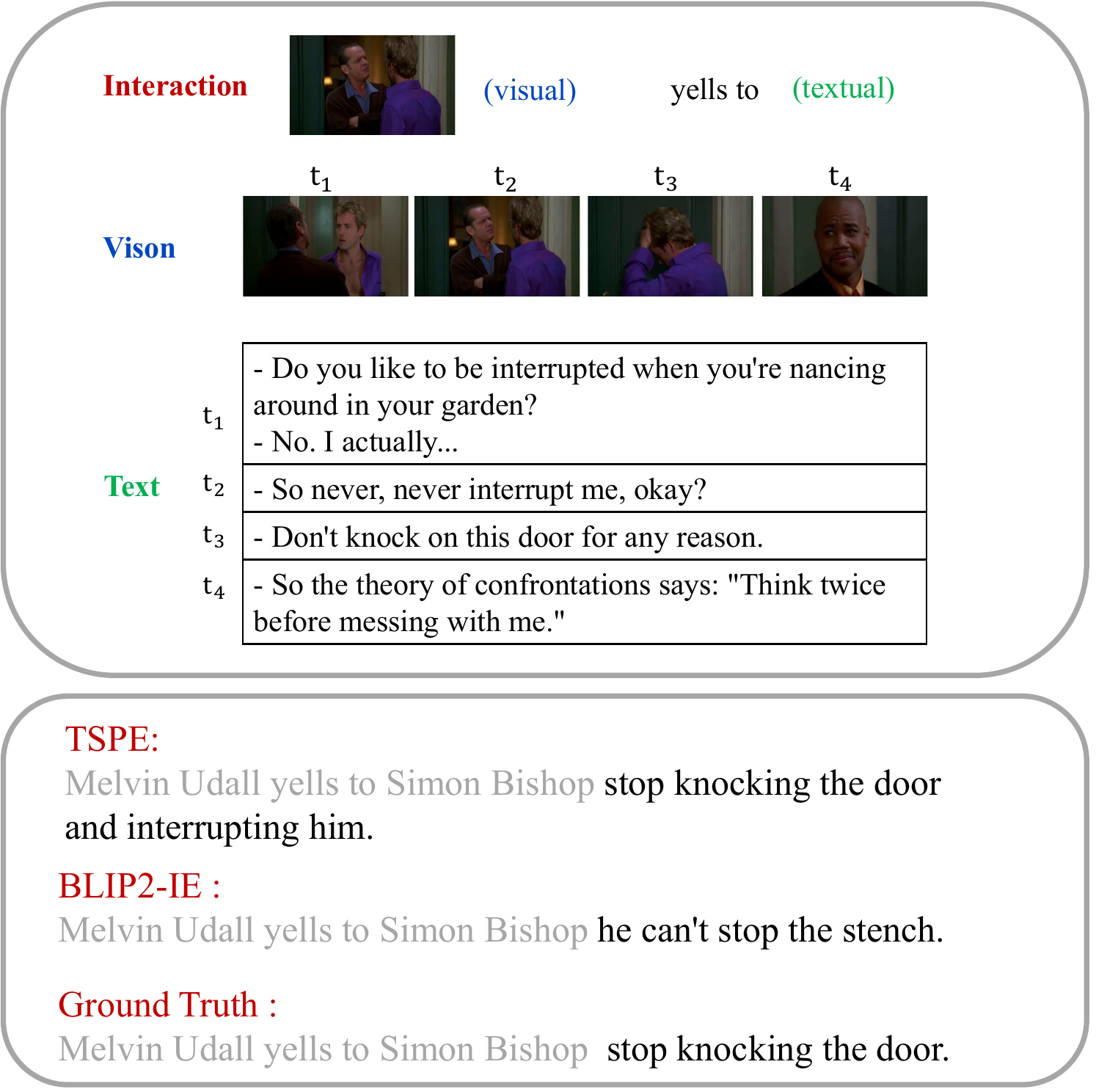}
    \caption{Local stage cases. A short video clip consists of synchronized frames and dialogues, and the interactions between characters. The gray fonts in the image represent that they are not involved in calculating metrics such as BLEU. Because the interaction text is already in the input, we will exclude the interaction part from the generated text when evaluating.
    }
    \label{fig:interaction}
\end{figure}

\begin{figure}[t]
    \centering
    \includegraphics[width=0.5\textwidth]{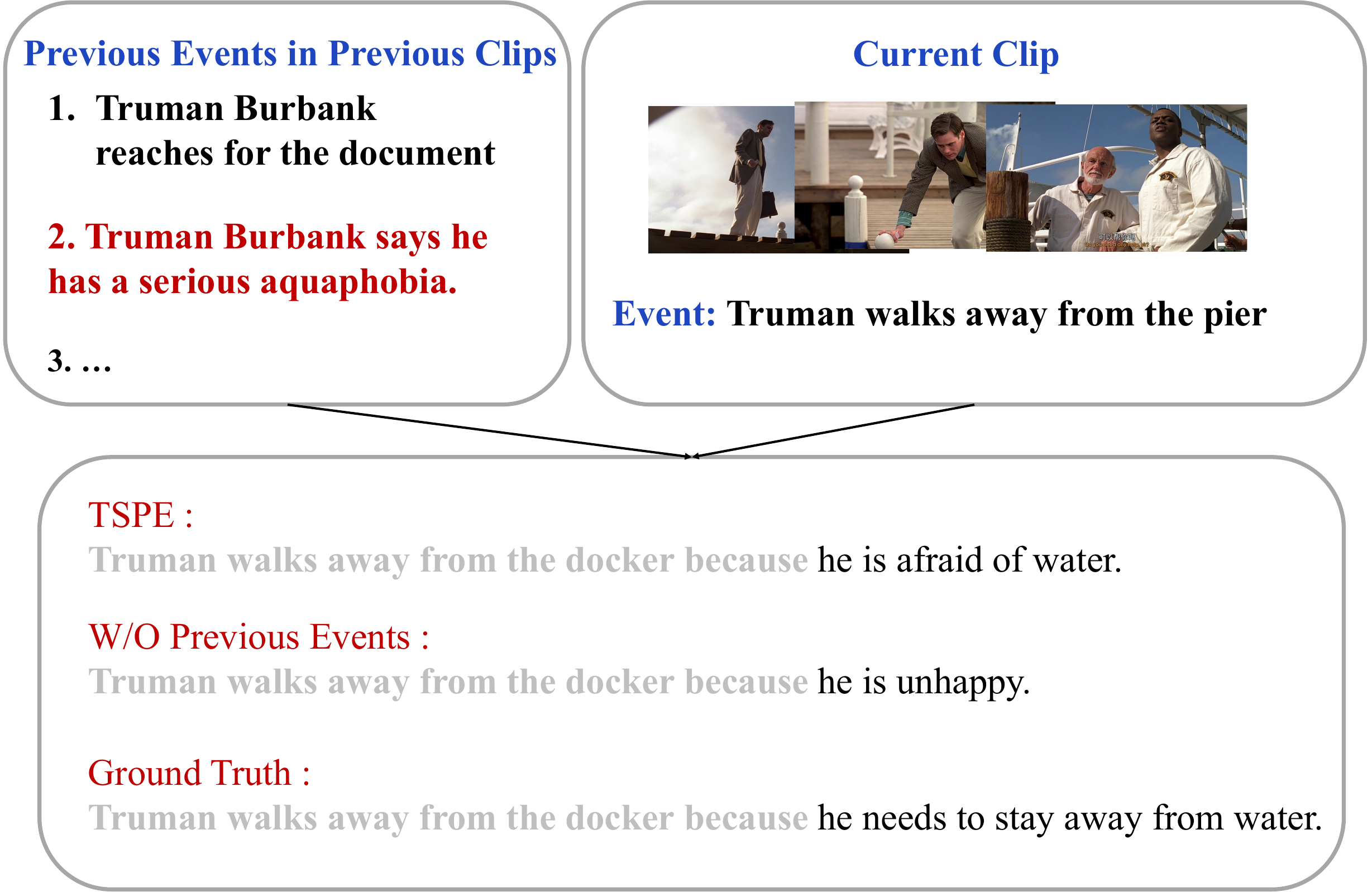}
    \caption{Global stage cases. We generate reasons behind events at the level of the whole video. We must consider the previous events in previous clips when generating reasons.}
    \label{fig:event}
\end{figure}

\subsection{Visualization}
\label{sec:case}
Figure \ref{fig:interaction} illustrates a video clip. Our goal is to generate the descriptions of events, given the interactions as known information. However, not all visual and textual information in the video clip is relevant to character interaction. For instance, in Figure \ref{fig:interaction}, the frame and dialogue at time $t_4$ are unrelated to the interaction "yells to". Our model can effectively filter out the irrelevant dialogue and frame and generate the description: to stop knocking on the door and interrupting him. In contrast, the baseline BLIP-2 model fails to do so and generates content that is not related to interaction.

To generate the reasons for the events, we considered the causal relationships between the events. Many events are caused by previous events, rather than just determined by the information in the current clip. For example, Figure \ref{fig:event} shows that Truman is afraid to approach the pier because of his hydrophobia. This reason cannot be directly derived from the current clip but requires reference to the previous clip, where Truman said he had a severe fear of water. Therefore, when we generate reasons, we combine the previous events and the current events. If there are no previous events, the generated reasons are not accurate enough.

\section{Conclusion}
This paper presents TSPE, a novel two-stage framework for generating attributions for story events in long videos. The local stage focuses on describing the events by using an interaction-aware prefix that estimates the relevance between multi-modal cues and social interactions. The global stage focuses on attributing the events by using an inferential knowledge graph that enhances the semantic similarity between related events, and an event-aware prefix that measures the logic correlations between events. We evaluate our framework on two real-world datasets and show that it achieves better performance than several SOTA methods.

\bibliographystyle{IEEEtran}
\bibliography{references}

\begin{IEEEbiography}
[{\includegraphics[width=1in,height=1.25in,clip,keepaspectratio]{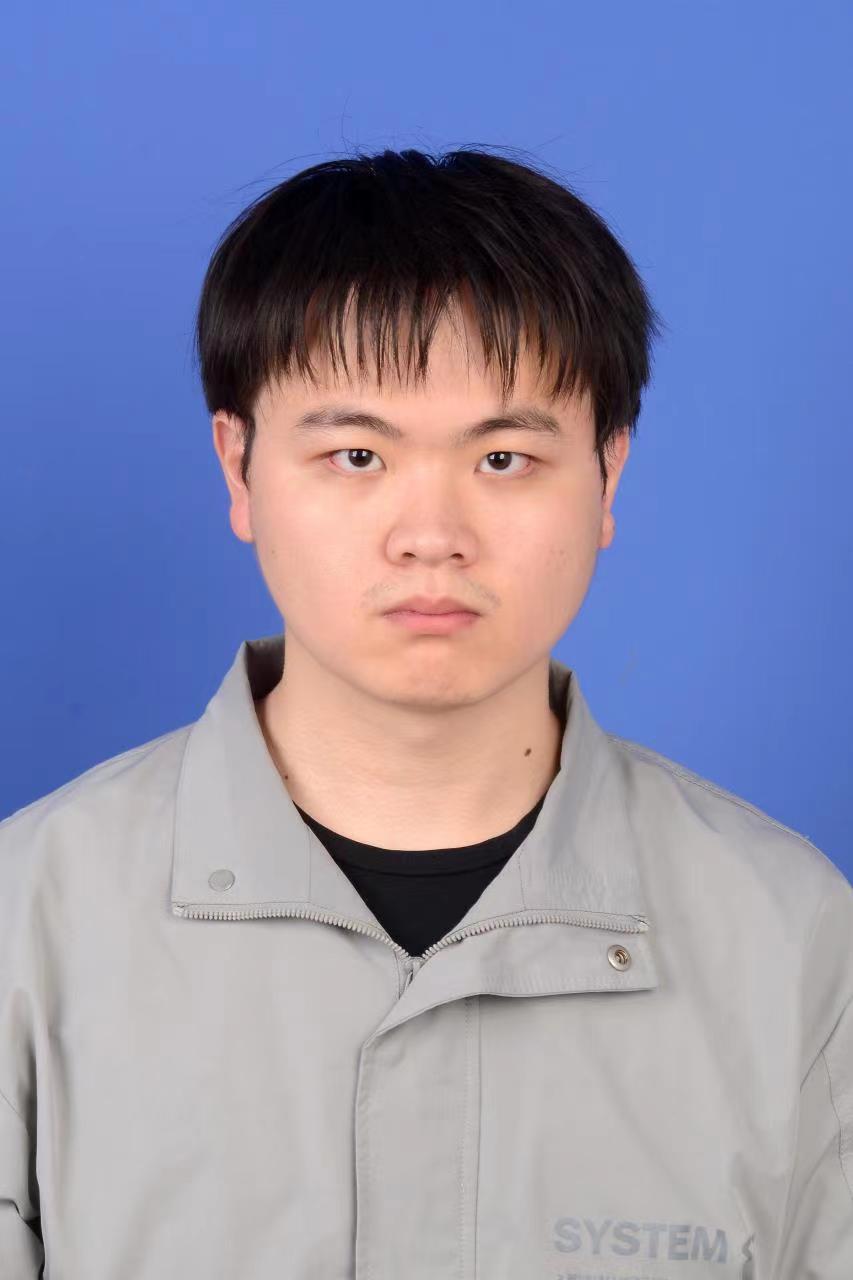}}]{Yuanjie Lyu} received the B.S. degree from the Tianjin University(TJU), Tianjin, China, in 2022. He is currently working towards the M.S. degree with University of Science and Technology of China(USTC), Hefei, China. He is working with Key Laboratory of Big Data Analysis and Application. His major research interests include Multi-modal Understanding and Nature Language Processing.
\end{IEEEbiography}
\begin{IEEEbiography}
[{\includegraphics[width=1in,height=1.25in,clip,keepaspectratio]{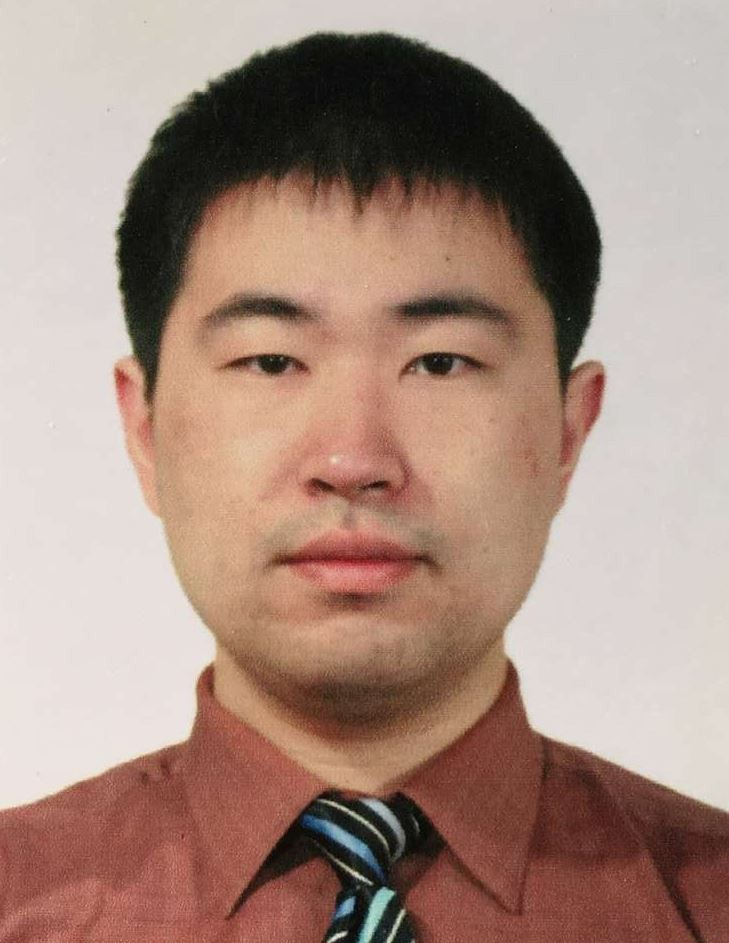}}]{Tong Xu}
 (Member, IEEE) is currently working as a Professor at University of Science and Technology of China (USTC), Hefei, China. He has authored more than 100 top-tier journal and conference papers in related fields, including TKDE, TMC, TMM, TOMM, KDD, SIGIR, WWW, ACM MM, etc. He was the recipient of the Best Paper Award of KSEM 2020, and the Area Chair Award for NLP Application Track of ACL 2023.
\end{IEEEbiography}

\begin{IEEEbiography}
[{\includegraphics[width=1in,height=1.25in,clip,keepaspectratio]{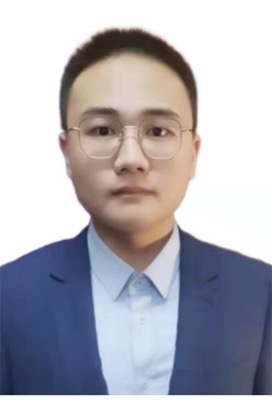}}]{Zihan Niu} received the B.S. degree from the University of Science and Technology of China(USTC), Hefei, China, in 2023. He is currently working towards the M.S. degree with USTC, Hefei, China. He is working with Key Laboratory of Big Data Analysis and Application. His major research interests include Multi-modal Understanding and Multi-modal Machine Learning.
\end{IEEEbiography}

\begin{IEEEbiography}
[{\includegraphics[width=1in,height=1.25in,clip,keepaspectratio]{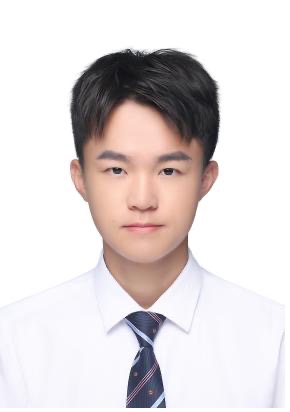}}]{Bo Peng} received the B.S. degree from the University of Science and Technology of China(USTC), Hefei, China, in 2023. He is currently working towards the M.S. degree with USTC, Hefei, China. He is working with Key Laboratory of Big Data Analysis and Application. His major research interests include Multimodal Understanding and Long Video Semantic Understanding.
\end{IEEEbiography}

\begin{IEEEbiography}
[{\includegraphics[width=1in,height=1.25in,clip,keepaspectratio]{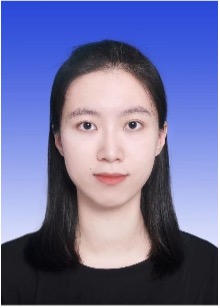}}]{Jing Ke} received the B.E. degree in Fuzhou University, Fuzhou, China, in 2023. She is currently working towards the M.A. degree with USTC, Hefei, China. She is working with Key Laboratory of Big Data Analysis and Application. Her major research interests include Multi-modal  Machine Learning and Disinformation Detection.
\end{IEEEbiography}
\begin{IEEEbiography}
[{\includegraphics[width=1in,height=1.25in,clip,keepaspectratio]{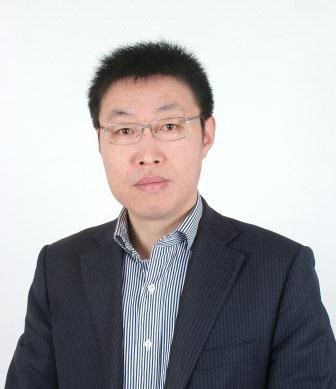}}]{Enhong Chen}
 (Fellow, IEEE) is a professor and vice dean of the School of Computer Science, USTC. His general area of research includes data mining and machine learning, social network analysis, and recommender systems. He has published more than 200 papers in refereed conferences and journals, including Nature Communications, IEEE/ACM Transactions, KDD, NIPS, IJCAI and AAAI, etc. He was on program committees of numerous conferences including KDD, ICDM, and SDM. He received the Best Application Paper Award on KDD-2008, the Best Research Paper Award on ICDM-2011, and the Best of SDM-2015. His research is supported by the National Science Foundation for Distinguished Young Scholars of China.
\end{IEEEbiography}

\end{document}